# A new Video Synopsis Based Approach Using Stereo Camera

Talha Dilber, Mehmet Serdar Güzel, Erkan Bostancı[*]

*Abstract*— In today's world, the amount of data produced in every field has increased at an unexpected level. In the face of increasing data, the importance of data processing has increased remarkably. Our resource topic is on the processing of video data, which has an important place in increasing data, and the production of summary videos. Within the scope of this resource, a new method for anomaly detection with object-based unsupervised learning has been developed while creating a video summary. By using this method, the video data is processed as pixels and the result is produced as a video segment. The process flow can be briefly summarized as follows. Objects on the video are detected according to their type, and then they are tracked. Then, the tracking history data of the objects are processed, and the classifier is trained with the object type. Thanks to this classifier, anomaly behavior of objects is detected. Video segments are determined by processing video moments containing anomaly behaviors. The video summary is created by extracting the detected video segments from the original video and combining them. The model we developed has been tested and verified separately for single camera and dual camera systems.

## I. INTRODUCTION

Today, rapidly developing technology has been included in every aspect of our lives, however, audio, visual and many other data have begun to accumulate. Big data has been expressed with different definitions by the pioneers of the software field in the recent past. The etymology of big data dates back to the mid-1990s. It was first used by former Chief Scientist John Mashey to refer to the processing and analysis of large data sets (Diebold, 2012). Simply put, big data specifically includes larger and more complex datasets than new data sources. These datasets are so bulky that traditional data processing software cannot manage them. But this large volume of data can be used to solve problems you couldn't handle before.

Among the data types that make up big data and also touch our daily life, camera and video images have reached a remarkable size (Tiejun Huang, 2014). Dozens or even hundreds of cameras are used for different purposes, from banks to museums, from ATMs to official government offices. For example, some cameras are located at the entrance and exit of residential areas and record vehicles passing through highways. Some cameras may also be positioned to see more pedestrians on the pavement on the streets. These cameras are mostly in high resolution, provide 24/7 service and at the same time all the data they produce is stored in certain places. For this reason, the amount of data produced is very large and data storage areas are actually filled with produced data. It is a known fact that most of this data is unnecessary. When a security camera image needs to be examined, the length of the image to be looked at and the possible number of different cameras that may occur are also taken into account and detecting the meaningful ones in the data can reach the limits that will force human potential. Summarizing, shortening, and eliminating unnecessary parts of the data will bring great benefits in many ways. First of all, we have a few questions. What are the unnecessary parts? How to identify important and important images that we need to keep? How to distinguish between unnecessary movements and important motion that must be save in a video with full of regular motion? These questions are the main problems of our research.

Various solutions are available for data reduction. But each solution has different advantages and disadvantages. One solution is to extract the summaries of the videos. This method can be in two ways. The first method is based on the principle of eliminating the unnecessary parts of the video before it is recorded in the main recorder and recording only the necessary parts in the main recorder. The second method is based on the analysis of the data produced in certain periods of the day and the week after the video recording is completed and the removal of unnecessary parts. There is a major problem involving both of these methods. This main problem is how to determine which part of the data is meaningful and which part is meaningless during abstraction.

In our research, I wanted to try the multi-object detector and tracker, which was developed using Convolutional Neural Network (CNN) as the algorithm design to be used in determining the areas where video summary should be extracted, with classification algorithms. A Multi object viewer will run over a video or image of a live camera. The multi-object tracker will send the types and movements of all detected objects to one classifier. At the same time, by sending the movements of the object detected by the multi-object tracker to the classifier, the results of which object types the motion made from the classifier belong to will be received and it will be checked whether the multi-object tracker acts in accordance with the type of the object detected. In the predictions made by the classifier on the detected motion, motion mismatch above a certain level will be detected. This method is an anomaly detection process. Parts of the video with motion inconsistency detected will be cut and added to the summary section. For this reason, routine works that do not deserve attention for each object type have been eliminated, and the movements that attract attention will be determined as the video summary part, separated from the other parts and combined in the summary section.

Here are our thoughts on the stereo camera video summarization approach. First of all, both cameras must have the same field of view, even if they have different angles. Video summary extraction will be applied in case of anomaly detected by 2 cameras using multiple object trackers and classifiers at the same time at the same time. Thus, anomalies detected by using multi-object tracker and classifier will be confirmed with two different viewing angles and will strengthen the possibility of anomalies.

## II. RELATED WORKS

Here is a brief summary of the literature review on video summarization:

Corresponding author:ebostanci@ankara.edu.tr

Video summarization applications and the first use of video summarization were in the media, TV broadcasting. A study was conducted to extract the video into a combined picture summary (Tao Chen, 2012). An application has been made for the summary extraction process by automatically creating the story of a movie.

In addition to the films, the video summary of the prominent parts of the sports competitions is a subject that has attracted the attention of scientists and is an area of commercial concern. In a study in 2011, it was aimed to summarize football matches in terms of the situation within the match (Hossam M. Zawbaa, 2011). Situations such as scoring a goal and using a penalty during the match were included in the summary and the research, which has a commercial aspect, was signed.

As video recorders have become easier for the average person to receive these days, it has increased the amount of video recorded by people in various places at any given time. In a study conducted in 2011, as a result of the analysis of daily videos, a video summarization research was conducted that provides efficient storage of videos as well as individual behavior analysis (Wei-Ting Peng, 2011). Videos are analyzed with rule-based algorithms.

According to a report published in 2015, more than 250 million cameras in professionally installed surveillance systems in 2015 are located in high-definition and closed networks. A significant amount of security personnel are assigned to watch these surveillance videos only. Instead, this painstaking work could have been accomplished by an intelligent video summarization system that provided a highlight for long surveillance videos. When a video summary of critical situations is produced and especially when multiple cameras are used to monitor a location, it will be very easy for security personnel to interpret. With multi-screen summaries, it facilitates quick browsing of important clips of videos. This problem is actually the focus of our resource. A study in this area was published in 2016 (Po Kong Lai, 2016). They use an algorithm over object detection.

An area that is closed circuit work and also has its own content is camera observations inside industrial buildings. It has become widespread to install cameras inside industrial buildings. In order to get information from these cameras and to ensure that the errors that have occurred are detected by Machine Learning techniques, a research group made a publication in 2011 (Athanasios S. Voulodimos, 2011). In this publication, they describe a Classification-based Video Summarization algorithm.

Video Summarization algorithms have also been developed in the medical world. He published a research on this subject in 2011 (Sourya Bhattacharyya, 2011). In this study, it was ensured that the seizure status of the babies was determined by cameras that specifically observed newborn babies. While one person is required to supervise 24 hours a day for each baby in the intensive care unit, thanks to the work done, it is possible to observe the seizure status of babies for 24 hours without interruption with the help of cameras that monitor babies.

The advent of wearable cameras on the market has allowed people to record videos from the person's eyes, which is called egocentric video. The ability to record egocentric videos has created scope for numerous computer vision challenges, ranging from sociobehavioral modeling to examining repetitive patterns in an individual's diary records. Since they are long videos in nature, egocentric videos must be summarized in order to efficiently analyze the objects and events appearing in front of the camera carrier. In a study conducted in 2015, a study was published on creating a video summary of daily activities from egocentric videos (Jia Xu, 2015). Among the problems they encounter are the detection of poor quality, unclear images, and the determination of the intention of the person who shoots the egocentric video. The scope of the study expands in line with these problems.

Self-centered videos are a godsend for people with short-term amnesia, as their activities are recorded throughout the day. But it's impossible to watch the log to find that part of the video that's interesting. A brief summary of these records would be of great help to people with these types of health conditions. In 2017, a study was published for patients with memory loss and similar disease (Rameswar Panda, 2017). In this study, daily self-centered videos are summarized and stored. A sick person is a subject, place, object, etc. When you want to get information about the subject, you are provided with the support of the search engine and the summary videos of the subject sought. It also provides access to the full video from which the summary video was cut.

The paper, published in 2011, proposes a new method for creating customized video summaries for application mainly to online videos (Dim P. Papadopoulos, 2011). The novelty of this approach is that the video summarization problem is considered as only getting one image per query. The method mentioned is that each frame is considered a separate image and will be explained by Compact Composite Descriptors (CCD) and a visual word histogram. The method used a powerful Self-Growing and Self-Organized Neural Gas (SGONG) network to classify the frames into clusters. Its main advantage is that it automatically adjusts the number and topology of generated neurons. Thus, after the training, SGONG is ensured to give us an appropriate number of output classes and centers. Extracting a representative keyframe from each set resulted in the creation of the video summary.

In an article published in 2014, the online summarization method was tried using the Gaussian mixture model (Shun-Hsing Ou, 2014). As a result of the experiments, it is said that the method used outperforms other online methods in both summarization quality and computational efficiency. It has been argued that hashing can be created with a shorter latency and much lower computational resource requirements.

In a publication made in 2017, content-based video retrieval, indexing and recommendation systems are described (Yashar Deldjoo, 2017). It is stated that their goal is to suggest items/videos to users based on the visual content of the videos.

A special class of video summary, movie synopsis, was studied in 2013 (Chia Ming Tsai, 2013). In the published article, a two-stage scene-based movie summarization method was developed based on mining the relationship between the role ensembles, since the role ensembles in the previous scenes are generally used to develop the role relationship in the following scenes. In the analysis phase, a social network was created to characterize interactions between role communities. In the summarization stage, it is proposed to identify viable summary combinations of scenes and select an information-rich summary from these candidates based on social power conservation.

A 2012 study compared the summarization methods used in large videos (Muhammad Ajmal, 2012). The case in which each method is most appropriate is discussed. Recommendations are provided to reduce the burden of processing unnecessary video frames and to help navigate very large videos quickly.

Another important application of video summation is its use in the analysis of medical videos such as endoscopic videos, recordings of medical procedures, diagnostic hysteroscopy videos, and so on. Medical videos vary greatly depending on the duration of medical procedures. Summing up long medical videos provides efficient analysis of procedures and helps teach procedures to medical students. In an article published in 2011, a study was conducted on bronchoscopy examinations (Mikołaj I. Leszczuk, 2011). When the whole procedure is made into a video, the important parts are summarized and similar cases are compared. In this way, the diagnosis, which takes 20 minutes, will be accelerated.

Professionals often need to search for similar cases, browse libraries containing many diagnostic hysteroscopy videos, or review a video of a particular case. Video search and browsing can be used in many situations, such as in case-based diagnosis where videos of previously diagnosed cases are compared, in case referrals, in reviewing patient records, and in support of medical research (eg, in human reproduction). In a paper published in 2011, a technique was proposed to facilitate the screening task, to estimate the clinical relevance of video segments in diagnostic hysteroscopy (Wilson Gavião, 2011). Basically, the proposed technique associates clinical relevance with the attention drawn by a diagnostic hysteroscopy video segment during video acquisition. For example, if an image has some waiting time, it means that the doctors there examine it and attract attention. In this study, general methods of summarizing videos and how they apply to diagnostic hysteroscopy videos are reviewed. They say that they concluded that the proposed method contributes to the field of hysteroscopy with a special summary and display method for video.

Summarizing news videos allows us to quickly get past the important patterns that appear in news videos. An article was written in this area in 2011 (Jie Ling Lai, 2011). They tried to reach the result by using known methods such as motion detection and feature extraction.

Unmanned aerial vehicle (UAV) and drone-based surveillance are gaining popularity day by day. Cameras are being integrated into new devices such as drones and robots, equipping them to record visual content in many places inaccessible to humans. Summarizing these videos will make this new category of videos easier to interpret. For this purpose, an article published in 2012 has been written (Athanasios S. Voulodimos, 2012). This article discusses a video summarization method that separates keyframes from industrial surveillance videos, thus significantly reducing the frame count without loss of meaningful semantic content. They proposed to use the generated summaries as a training set for neural network-based Evaluative Rectification. Evaluation Validation is a method that uses expert user feedback on the accuracy of the event recognition framework on a portion of the data to improve future classification results.

In the literature review, video summary applications and research for many different purposes were examined. In general, it was tried to examine his articles aiming to extract a video summary with different purposes, not the same style. It has been seen that video summaries are used in many other fields, and the above studies are a very small part of all the literature in the field of video summation.

## III. METHODS

### A. Multi Object Detection And Tracking

In our work that we developed for the video summary, the YOLO image processing tool, which has been developed using Convolutional Neural Network, which has good references recently, was used. It is an algorithm used to detect objects in an image. As the name suggests (YOLO – You Only Look Once), it looks at the whole image only once, passes through the neural network once and detects objects (Sujata Chaudhari, 2020).

The YOLO architecture was developed using and inspired by F-CNN (fully convolutional neural network). When processing an image, the processed image is passed over F-CNN only once and the output is estimated. After the object is detected, it detects the most appropriate object type and size by processing the probabilities of object types and object sizes in all determined probabilities in order to confirm the size and type of the object, as if solving a regression problem. The equivalent of the detected object size can be displayed on the screen as a frame. The said object size is only related to the area it covers on the screen. A clear size measurement cannot be made. In a single evaluation, a single neural network predicts object sizes and class probabilities directly from full-size images. In most image processing techniques, object detection is made by processing all possibilities of different sizes in the image. This increases the burden of object detection many times over.

### B. Classfication

There are many classification algorithms used today, but it is not possible to conclude which one is superior to the other. Classification algorithms have their own pros and cons, or we can say that classification algorithms have characters. For this reason, which classification algorithm we will use depends on the application and nature of the dataset. For example, linear classifiers such as logistic regression may outperform sophisticated models such as Fisher's linear discriminant if the classes are linearly separable.

Not all classification algorithms supported the features we needed in our research. These features were lazy learning and partially learning. The reason we needed these features was that we wanted to do real-time learning by accumulation within the video. For this reason, we used the Stochastic Gradient Descent Classifier algorithm in our research.

### C. Anomaly Detection

Anomaly detection is the work of finding data that differs greatly from what has been seen before during a data observation. It is one of the main problems of artificial intelligence. Anomaly detection in the functioning of our research is one of our key tasks. Because, for the places where the Video Summary will be extracted, situations with anomalies will be selected.

Detecting anomalies in data is an important capability for humans and artificial intelligence. People often want to detect anomalies to detect early signs of danger or to anticipate opportunities. Anomaly detection systems are used by artificial intelligence to detect credit card fraud. In cybersecurity, anomaly detection is used to detect intrusion, warn industrial equipment to be predicted and maintained, and discover attractive opportunities in the stock market. The typical anomaly detection method is a single-class classification task where the target's data is classified as normal or abnormal.

There are different possible scenarios for anomaly detection methods. In supervised anomaly detection, training examples regarding normal and abnormal patterns are given for training purposes before anomaly detection is performed. In this way, anomaly detection tools developed with pre-training can work better than those without training. But it is not always possible to be able to teach. In order for the training to be available, you must know what anomalies you may encounter. If you are not likely to know the anomalies you may encounter; For example, since the firewall used in cyber

Recently, we observe that classification-based methods produce quite good results on anomaly detection. In classification-based methods, semi-supervised scenarios can also be run.

IV. VIDEO SYNOPSIS SYSTEM

Python programming language was used in the software project of our research. The main reason for this is the Python programming language is used in many artificial intelligence applications used today. It provides extensive library facilities.

Computer vision was performed with artificial neural networks that detect objects of the YOLO tool, which was previously trained using Python programming language and OpenCV library. The information of the detected objects is normalized and sent to the classifier. According to the previously known type of the object, it is determined whether security cannot be controlled samples of unknown computer viruses, we should be able to detect anomalies without supervised training. It is completely unsupervised anomaly detection that can be used at the point we are talking about. It attempts to detect abnormal data in a data stream containing normal and abnormal data.

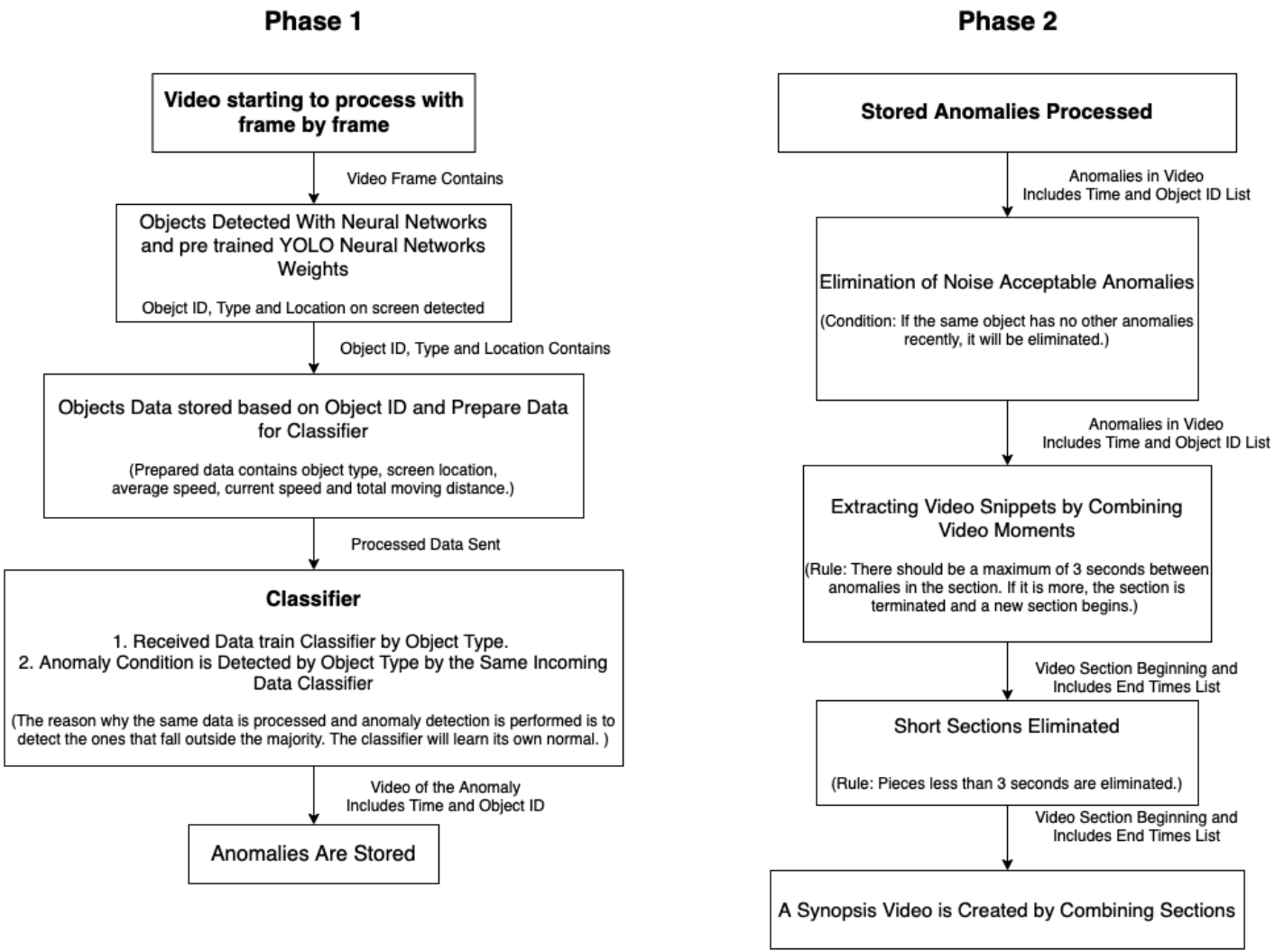

Fig. 1. Video Synopsis General Algorithm

it contains anomalies from the classifier. It is sent to the classifier as training data regardless of the anomaly status. A summary process is created by passing through various filters and processes for video segments with anomaly detection.

Algorithms used on this flow; Trained artificial neural networks are used for image detection and classification algorithm is used for anomaly detection. The rules with solid values are given as an example in the flow chart and may vary.

### A. Goal and Motivation

The aim of our research is to detect anomaly on video images taken with fixed cameras. For anomaly detection, there is a tool called YOLO, which detects and tracks objects. As soon as the detected objects enter the camera, all their movements are recorded and converted into a data type suitable for sending to the classifier. The classifier is trained with the transformed data. Then the classifier is asked how much the object belongs to its class. For example, a car object is recognized and followed. The classifier is asked how well this car behaves like a car. If the classifier says it is not behaving like a car, then an anomaly is assumed to exist.

The classifier is both trained and used in the classification query. This may seem like a contradiction. However, since the classifier is subjected to continuous training in the long term, it becomes able to distinguish between those who behave differently and those who do not. It can be thought of as similar to the rule "Students who fall under the bell curve fail in class".

### B. Preparing the Video

The reason for the preparation part of the video is that the parts of a summarized video cannot be displayed only in a combined form. The summary of the video should contain some information about the original video that was created. It is very important to know which second of the original video is in the summary video. It was thought that it was important to display the anomaly detection, which is the basis of our resource, in the summary video, and a study was carried out on these subjects. As part of these studies, a time indicator was added to the upper left part of the video and object movements with anomalies were marked.

### C. Identifying Video Summary Parts

Determining the video summary parts is the most important stage of the video summary extraction process. Almost all algorithmic operations are performed at this stage. In this phase, 2 technologies were used and then various algorithms were developed and this phase was completed.

#### 1) Using YOLO

YOLO (You Only Look Once) is a machine vision tool with object recognition and tracking capabilities developed with convolutional neural networks. The main difference in its development is that while processing an image, not all sub image parts in the image are processed, but a whole image part is processed only once. In many machine vision algorithms, an image is processed by looking at its subparts (Yılmaz et. al, 2020).

As it is known, convolutional neural networks are machine vision algorithms that are suitable for processing with GPU for this reason, which can be very slow when run with a CPU that requires a lot of processing. The YOLO tool used in the resource work needs to be run with a GPU to avoid being slow. Therefore, at least one GPU and CUDA library must be installed on your computer that you need to run in our study. In the YOLO tool, the CUDA supported Open-CV library is used in the python programming language. Specifically, CuDNN tools of the CUDA library are used.

In the work we have done for our resource, the YOLO tool is our reference point that makes the first transaction. When the frame of a video is received, this frame is first sent to the YOLO program. The YOLO tool outputs a list. Each element in this list holds the following data:

• Identification number of the object assigned by YOLO (It is created with unique number content)

• Type of object (car, human, truck, dog etc. includes a total of 86 object types)

• Position of the object on the screen (horizontal and vertical position in pixels)

• The width of the object on the screen (width and height in pixels)

• Probability of matching with the type of object (output is produced between 0 and 1. Objects are defined by placing a limit on the probability of fit. For example, results below 0.5 are ignored.)

This list from YOLO is sent to the classifier step for processing.

#### 2) Using Classifier

The reason why we wanted to use a classifier in our study is that it is a structure that learns its own truth instead of thinking about the possibilities for each situation while processing a video. For example, we want to detect vehicles going in the opposite direction while examining the images of a camera observing a road. Normally, we define the direction of the solution path that comes to our mind first in our work and we can detect who is going in the opposite direction. This solution will only be valid for one path and one time interval. That is, when we put the camera with the same program on another road with a view angle facing another direction, the program will not work properly. Or when there is a modification on the road, the program will not work properly when the road is double lane. In our study, instead of trying such rote solutions, we tried to produce a solution that would cover all possibilities.

In this solution, which will cover all possibilities, the task of the classifier is key. The classifier tries to classify the detected and tracked objects with the YOLO tool in a fixed camera view. In other words, the motion of this object is taught to the classifier while the type of the object is already known. The same classifier is then asked if this object is moving correctly relative to its other congeners. If the classifier finds the movements of the object objectionable, this situation is taken into the anomaly category and stored for summarization. For example, let's say there are cars going in one direction on a road. The information of cars going in one direction is constantly taught to the classifier. If a vehicle enters from the opposite direction, the classifier detects a very different movement in the car behavior it has learned so far and detects an anomaly.

In our research, the most important indispensable criterion while choosing the classifier has been piecemeal learning. Piecemeal learning is when an existing working classification system receives new teaching data, it accepts this data and classifies it by taking this data into account in its subsequent operations. We have a limited set of classification algorithms left, as there is a feature that is not suitable for every classification algorithm.

Data from the YOLO tool needs to be transformed for classification. A structure was considered for this transformation. First of all, the location of the detected object was important as well as the historical data. For this reason, it was tried to create a training data based on the past by storing the data received from the YOLO tool with the object ID. For example, it is not normal for a person or a car to wait for a long time in the middle of the road. For this, the following data has been prepared for training and is sent to the classification unit as the following 5 variables:

1. Center (X) of the object relative to the horizontal axis of the screen. The result is returned in pixels.

2. Center (Y) of the object relative to the vertical axis of the screen. The result is returned in pixels.

3. The current speed of the object (Distance between the center of the previous video frame and the current center)

$$\sqrt{(X[n] - X[n-1])^2 + (Y[n] - Y[n-1])^2}$$

(2-1)

4. Total displacement of the object from start to finish

$$\sqrt{(X[n] - X[0])^2 + (Y[n] - Y[0])^2}$$

(2-2)

5. Average velocity of the object from the start

$$\frac{\left(\sqrt{(X[1]-X[0])^2+(Y[1]-Y[0])^2}+\cdots+\sqrt{(X[n]-X[n-1])^2+(Y[n]-Y[n-1])^2}\right)}{n}$$

(2-3)

In the formulas 2-1, 2-2, and 2-3, the variable n indicates the amount of video frame in which the detected object is followed. X[0] indicates the horizontal position of the object at the first instant, X[n] the horizontal position of the object at the last instant, Y[0] the vertical position of the object at the first instant, Y[n] the vertical position of the object at the last instant.

As you can see above, it has been tried to detect long-term remarkable situations by ensuring that the data sent to the classifier establishes a link with the past movements.

Training data is not sent to the classifier one by one for each object detection. Because this situation can adversely affect the performance of the classifier. For this reason, the learning methods of the classifier are not called without around 100 object definitions.

As we explained in our research, the system performs a self-learning process. According to each object type, general movements will be learned and non-general movements will be detected. But there is something that cannot be overlooked here. His predictions will not be correct before he reaches a certain maturity as classifier learning. For this reason, the limits of asking the classifier were created. In our experiments, in general, the results of the classifier were not taken into account without learning 200 pieces of an object type.

One of the main problems of many projects dealing with artificial intelligence today is to determine the boundaries at the optimum level. In our study, there are limits to which the optimum should be investigated. The most important is the limit of the value given by the classifier whether there is an anomaly or not. Another limit is the acceptance limit of the specified accuracy rates of the objects detected by the YOLO tool. In other words, the YOLO tool can mistake a garbage can for a human in a picture with a probability of 0.1 (a value between 0 and 1). In our research, a limit value is determined for these possibilities. Another limit relates to how many training data the classifier has given the answers taken into account. In our experiments, this limit was used as 400. In other words, after 400 training data, anomaly detection was started with the classification algorithm. The final limit value is the number of data that should be sent to the classification algorithm for training. In our experiments, this limit value was mostly used as 100.

*D. Processing of Anomalies*

The processing of anomalies has matured as a result of our observations after experimental studies. Because it has been observed that there are anomalies that can be called "noise", which is the image processing term, within the anomalies. For example, a car is on the screen for 10 seconds, in its normal course. During this observation, 1 or 2 anomalies are detected, but there is no other anomaly related to the same object that supports or follows these anomalies. In these cases, a summary of these anomalies should not be created. Because half-second video summary pieces can be created. In order to avoid this undesirable situation, after the video anomaly analysis was completed, all detected anomalies were grouped based on object ID. In this grouping, if there are objects with anomaly detected below a certain amount, these anomalies are removed from the list. For example, objects with less than 5 anomalies have been removed from the list. If there are anomalies that are far from each other by visiting the anomalies of each object one by one among the remaining objects, these anomalies were deleted. For example, an object has 15 anomalies and 13 of them are detected in a row at intervals of less than 1 second. However, 2 of them were detected far from the others. In these cases, the 2 we mentioned are deleted from the anomaly list. It is ensured that all anomalies are close to each other and repetitive on an object basis.

After the noisy parts of the detected anomalies are cleaned, these anomalies should be turned into video parts. Because if only one video frame is extracted for each anomaly and these frames are combined, the basic structure of the video will be disrupted and it may become incomprehensible. Developed a simple algorithm to identify video segments of detected and extracted anomalies.

1. Grouped anomalies based on object type will be collected in a single list.

2. The created list will be reordered according to the detection second.

3. Elements in this list will be looked at one by one.

4. If there is more than 1 second difference between the previous element's time and the current element's time, the video will have a section transition. The time of the current element will be the start second of the next segment.

Another situation that is not written in the algorithm to avoid confusion is the following. If there are less than 5 anomalies in a video segment, this segment is canceled. Because in this section, anomaly detection has been made as much as we can count noise.

*E. Creating Video Summary*

When creating a video summary, an original video is first reproduced to show anomaly frames and video seconds. Summary extraction process is carried out by processing a video produced this again.

The process of extracting the video summary is quite simple. Each element in a list represents a part of the video, and these elements must be sorted in the order in which they were processed. The elements in the list include the start and end seconds of the video track. Our video extraction feature takes the list we mentioned, reads the elements one by one, cuts the video segment between the seconds specified for each element and saves it to the file address we want as the result video. The file directory and name that should be saved are specifically specified in the configuration files.

Another problem to be encountered while extracting the video summary is this. When the differences between the video segments are shortened, for example, when 3-4 video segments are combined at intervals of 2 seconds, an annoying image occurs. A solution has been developed to avoid such inconveniences. For example, it must be combined within proximity of sections of video for less than 3 seconds. With this development, we started to offer a smoother video summary experience.

*F. Creating a Video Summary with Stereo Cameras*

In our study, the most important problem we encountered while creating a video summary with stereo cameras is how the video segments detected by the two cameras separately should be turned into a single video. The following solution has been accepted on this issue. Cases detected by both cameras at the same time will be included in both summaries, if only one camera detects an anomaly, it will be invalidated.

The solution we mentioned was put into code as follows. Two videos of two cameras were selected and given to our application. Anomalies and video segments of two videos were extracted. Video summarization was performed by taking the intersection sets of the sections of the two videos.

Creating a video summary on a stereo camera has an advantage over our method. If irrelevant anomalies within the scope of noise are encountered in both cameras, the probability of encountering will be considerably reduced and more stable results will be provided.

There is another problem with video summation with stereo cameras. This problem; If a video did not start at the same time as the other video, how are these two videos synchronized? As a solution to this problem, the video time difference input section was placed next to the videos given as input in our study. By specifying time differences between videos, other processes are not affected.

## V. RESULTS

In the experiments of our research, 20 experiments were carried out with a single camera, 2 experiments were carried out with a double camera. In single camera experiments, the variables used in our study were tested with 5 different videos. Results were produced by keeping the variables constant in stereo camera experiments.

In our stereo camera experiments, we need to have fixed cameras with the same field of view that shot at the same time. Since video data with these conditions could not be found, similar conditions were tried to be produced. Extracted different but intersecting segments of a single video. In this way, it was desired to obtain a side-by-side camera image facing the same direction. The same video was used in both single-camera experiments and stereo-camera experiments, providing a controlled experiment.

In all our experiments, SGD (Stochastic Gradient Descent) was used as the classification algorithm and the "yolov3-tiny" study in the YOLO tool.

*Table 2. Comparison of Single Camera Experiment Results*

| Video No / Scenario No | Experiment Values | | | | Experiment Results | | | |
|---|---|---|---|---|---|---|---|---|
| | YOLO Thresh. | Class. Thresh. | N. A. Thresh. | S. A. Thresh. | Total Sum. Time | Num. of Piece | Avg. Piece Time | Sum. Rate |
| | 0-1 | 0-1 | Piece | Piece | min. | Piece | sec. | % |
| 1.1 | 0.5 | 0.5 | 5 | 3 | 3:29 | 34 | 6.1 | 17.4 |
| 1.2 | 0.3 | 0.5 | 5 | 3 | 4:54 | 46 | 6.3 | 24.5 |
| 1.3 | 0.5 | 0.8 | 5 | 3 | 3:29 | 34 | 6.1 | 17.4 |
| 1.4 | 0.5 | 0.5 | 10 | 6 | 2:53 | 21 | 5 | 14.4 |
| 2.1 | 0.5 | 0.5 | 5 | 3 | 3:13 | 61 | 3.1 | 14.8 |
| 2.2 | 0.3 | 0.5 | 5 | 3 | 8:25 | 75 | 6.7 | 38.8 |
| 2.3 | 0.5 | 0.8 | 5 | 3 | 3:29 | 59 | 3.3 | 15 |
| 2.4 | 0.5 | 0.5 | 10 | 6 | 2:12 | 46 | 2.8 | 10.1 |
| 3.1 | 0.5 | 0.5 | 5 | 3 | 1:05 | 32 | 2 | 4.8 |
| 3.2 | 0.3 | 0.5 | 5 | 3 | 2:04 | 49 | 2.5 | 9.1 |
| 3.3 | 0.5 | 0.8 | 5 | 3 | 0:25 | 12 | 2 | 1.8 |
| 3.4 | 0.5 | 0.5 | 10 | 6 | 0:31 | 16 | 2 | 2.2 |
| 4.1 | 0.5 | 0.5 | 5 | 3 | 32:31 | 107 | 18.2 | 51.4 |
| 4.2 | 0.3 | 0.5 | 5 | 3 | 45:13 | 46 | 34.6 | 75.5 |
| 4.3 | 0.5 | 0.8 | 5 | 3 | 29:09 | 34 | 15.9 | 48.3 |
| 4.4 | 0.5 | 0.5 | 10 | 6 | 26:23 | 21 | 14.5 | 43.4 |
| 5.1 | 0.5 | 0.5 | 5 | 3 | 11:52 | 105 | 6.7 | 18.4 |
| 5.2 | 0.3 | 0.5 | 5 | 3 | 23:14 | 149 | 9.3 | 35.3 |
| 5.3 | 0.5 | 0.8 | 5 | 3 | 11:40 | 101 | 6.9 | 16.9 |
| 5.4 | 0.5 | 0.5 | 10 | 6 | 11:01 | 99 | 6.6 | 16.9 |

*Table 1. Comparison of Single-Stereo Camera Experiment Results*

| Video No / Scenario No | Experiment Values | | | | Experiment Results | | | |
|---|---|---|---|---|---|---|---|---|
| | YOLO Thresh. | Class. Thresh. | N. A. Thresh. | S. A. Thresh. | Total Sum. Time | Num. of Piece | Avg. Piece Time | Sum. Rate |
| | 0-1 | 0-1 | Piece | Piece | min. | Piece | sec. | % |
| 1.1 | 0.5 | 0.5 | 5 | 3 | 3:29 | 34 | 6.1 | 17.4 |
| 6 (Stereo) | 0.5 | 0.5 | 5 | 3 | 1:11 | 27 | 2.6 | 5.5 |
| 5.1 | 0.5 | 0.5 | 5 | 3 | 11:40 | 101 | 6.9 | 16.9 |
| 7 (Stereo) | 0.5 | 0.5 | 5 | 3 | 10:32 | 149 | 4.2 | 16.1 |

The title descriptions of the Experimental Values section in Table 1 and Table 2 are as follows:

• YOLO Thresh.: Video YOLO object detection limit.

- Class. Thresh.: Video Classification algorithm anomaly detection limit.
- N. A. Thresh.: The minimum number of anomalies that an object can have in the video.
- S. A. Thresh.: The minimum number of sequential anomalies that an object can have in the video.

## VI. CONCLUSION

Our study aims to bring a different perspective to the field of video summarization. In this context, a method that can unsupervised self-training during video processing, includes object-based learning, and uses anomaly detection with classification algorithm after unsupervised training has been developed. Using this method, video segments are determined by processing the video, which is a pile of pixels, and then video segments are extracted from the original video and a video summary is created as a combination of video segments.

The system we propose within the scope of our research can work with images taken with a fixed camera. Video summarization in the stereo camera system, which is the main purpose of our research, actually addresses limited areas. For example, the stereo camera system is available for cameras facing the same road or the same area.

In our single-camera experiments, the limitations we used in the studies were tested and observed. In the stereo camera experiments, a single camera performance comparison of the two cameras was achieved. As it is known, it is very possible that the results of an unsupervised learning system will contain uncertainty. When the relevant experimental results of this study were examined, it was observed that this situation prevailed. The most important point that attracted our attention in our experiments is the consistency of the results. The effect of the variables we gave in our experiments shows a consistency. In experiments on similar videos, the extracted video summary ratio is similar. In addition, the effect of different variables in the experiments on similar videos is very close to the video summary ratio result. The interpretation we made by looking at these coherences and similarities; if you expect a certain ratio of video summary, you can change the variables of our study in accordance with that ratio and the result can be as close as you want.

Some data were produced within the scope of the experiments, but it was not possible to perform real tests in our research. Because for an accurate experimental setup, an accident-like situation must occur after a camera has recorded for a long time. Long-term experiments should be repeated until the study matures. Experiments have been made on the inputs that we can access, since experimental work at this level is beyond research possibilities.

Video summarization is a field that has been studied extensively by scientists in recent years. Within the scope of this research, a new video summary processing approach was developed with the stereo camera system over the object-based unsupervised education model, contributing to the relevant literature.